%% file: gclr-aaai-2024.tex
\documentclass[letterpaper]{article} 
\usepackage[submission]{aaai24}  
\usepackage{times}  
\usepackage{helvet}  
\usepackage{courier}  
\usepackage[hyphens]{url}  
\usepackage{graphicx} 
\usepackage{natbib}  
\usepackage{caption} 
\usepackage{algorithm}
\usepackage{algorithmic}
\usepackage[utf8]{inputenc} 
\usepackage[T1]{fontenc}    
\usepackage{url}            
\usepackage{booktabs}       
\usepackage{amsfonts}       
\usepackage{nicefrac}       
\usepackage{microtype}      
\usepackage{xcolor}         
\usepackage{amsmath, amsthm}
\usepackage{multirow}
\usepackage{adjustbox, lipsum}

\usepackage{newfloat}
\usepackage{listings}
\DeclareCaptionStyle{ruled}{labelfont=normalfont,labelsep=colon,strut=off} 
\floatstyle{ruled}
\newfloat{listing}{tb}{lst}{}
\floatname{listing}{Listing}
\usepackage{tikz}
\usepackage{tikzpeople}
\usetikzlibrary{fit,arrows.meta,mindmap,shadows,backgrounds,calc,positioning,shapes.misc}
\usepackage{xcolor}
\usetikzlibrary{arrows,3d,calc}
\usepackage{tikz-3dplot}
\makeatletter
\def\tikz@lib@cuboid@get#1{\pgfkeysvalueof{/tikz/cuboid/#1}}

\def\tikz@lib@cuboid@setup{%
   \pgfmathsetlengthmacro{\vxx}%
      {\tikz@lib@cuboid@get{xscale}*cos(\tikz@lib@cuboid@get{xangle})*1cm}
   \pgfmathsetlengthmacro{\vxy}%
      {\tikz@lib@cuboid@get{xscale}*sin(\tikz@lib@cuboid@get{xangle})*1cm}
   \pgfmathsetlengthmacro{\vyx}%
      {\tikz@lib@cuboid@get{yscale}*cos(\tikz@lib@cuboid@get{yangle})*1cm}
   \pgfmathsetlengthmacro{\vyy}%
      {\tikz@lib@cuboid@get{yscale}*sin(\tikz@lib@cuboid@get{yangle})*1cm}
   \pgfmathsetlengthmacro{\vzx}%
      {\tikz@lib@cuboid@get{zscale}*cos(\tikz@lib@cuboid@get{zangle})*1cm}
   \pgfmathsetlengthmacro{\vzy}%
      {\tikz@lib@cuboid@get{zscale}*sin(\tikz@lib@cuboid@get{zangle})*1cm}
}

\def\tikz@lib@cuboid@draw#1--#2--#3\pgf@stop{%
    \begin{scope}[join=bevel,x={(\vxx,\vxy)},y={(\vyx,\vyy)},z={(\vzx,\vzy)}]
       \begin{scope}[canvas is yz plane at x=#1]
          \draw[cuboid/all faces,cuboid/edges,cuboid/right face] 
                (0,0) -- ++(#2,0) -- ++(0,-#3) -- ++(-#2,0) -- cycle;
          \draw[cuboid/all grids,cuboid/right grid] (0,0) grid (#2,-#3);
       \end{scope}
       \begin{scope}[canvas is xy plane at z=0]
          \draw[cuboid/all faces,cuboid/edges,cuboid/front face] 
                (0,0) -- ++(#1,0) --  ++(0,#2) -- ++(-#1,0) -- cycle;
          \draw[cuboid/all grids,cuboid/front grid] (0,0) grid (#1,#2);
       \end{scope}
       \begin{scope}[canvas is xz plane at y=#2]
          \draw[cuboid/all faces,cuboid/edges,cuboid/top face] 
                (0,0) -- ++(#1,0) --  ++(0,-#3) -- ++(-#1,0) -- cycle;
          \draw[cuboid/all grids,cuboid/top grid] (0,0) grid (#1,-#3);
       \end{scope}
       \draw[cuboid/hidden edges] (0,#2,-#3) -- (0,0,-#3) -- (0,0,0) 
                (0,0,-#3) -- ++(#1,0,0);
       \begin{scope}[canvas is yz plane at x=#1]
          \draw[cuboid/all faces,cuboid/right face,cuboid/edges,fill opacity=0] 
                (0,0) -- ++(#2,0) -- ++(0,-#3) -- ++(-#2,0) -- cycle;
       \end{scope}
       \begin{scope}[canvas is xy plane at z=0]
          \draw[cuboid/all faces,cuboid/front face,cuboid/edges,fill opacity=0] 
                (0,0) -- ++(#1,0) --  ++(0,#2) -- ++(-#1,0) -- cycle;
       \end{scope}
       \begin{scope}[canvas is xz plane at y=#2]
          \draw[cuboid/all faces,cuboid/top face,cuboid/edges,fill opacity=0] 
                (0,0) -- ++(#1,0) --  ++(0,-#3) -- ++(-#1,0) -- cycle;
       \end{scope}
       \path (0,#2,0) coordinate (-left top front)
                      coordinate (-left front top)
                      coordinate (-top left front)
                      coordinate (-top front left)
                      coordinate (-front top left)
                      coordinate (-front left top);
       \path (0,#2,-#3) coordinate (-left top rear)
                        coordinate (-left rear top)
                        coordinate (-top left rear)
                        coordinate (-top rear left)
                        coordinate (-rear top left)
                        coordinate (-rear left top);
       \path (0,0,-#3) coordinate (-left bottom rear)
                       coordinate (-left rear bottom)
                       coordinate (-bottom left rear)
                       coordinate (-bottom rear left)
                       coordinate (-rear bottom left)
                       coordinate (-rear left bottom);
       \path (0,0,0) coordinate (-left bottom front)
                     coordinate (-left front bottom)
                     coordinate (-bottom left front)
                     coordinate (-bottom front left)
                     coordinate (-front bottom left)
                     coordinate (-front left bottom);
       \path (#1,#2,0) coordinate (-right top front)
                       coordinate (-right front top)
                       coordinate (-top right front)
                       coordinate (-top front right)
                       coordinate (-front top right)
                       coordinate (-front right top);
       \path (#1,#2,-#3) coordinate (-right top rear)
                         coordinate (-right rear top)
                         coordinate (-top right rear)
                         coordinate (-top rear right)
                         coordinate (-rear top right)
                         coordinate (-rear right top);
       \path (#1,0,-#3) coordinate (-right bottom rear)
                        coordinate (-right rear bottom)
                        coordinate (-bottom right rear)
                        coordinate (-bottom rear right)
                        coordinate (-rear bottom right)
                        coordinate (-rear right bottom);
       \path (#1,0,0) coordinate (-right bottom front)
                      coordinate (-right front bottom)
                      coordinate (-bottom right front)
                      coordinate (-bottom front right)
                      coordinate (-front bottom right)
                      coordinate (-front right bottom);
       \coordinate (-left center) at (0,.5*#2,-.5*#3);
       \coordinate (-right center) at (#1,.5*#2,-.5*#3);
       \coordinate (-top center) at (.5*#1,#2,-.5*#3);
       \coordinate (-bottom center) at (.5*#1,0,-.5*#3);
       \coordinate (-front center) at (.5*#1,.5*#2,0);
       \coordinate (-rear center) at (.5*#1,.5*#2,-#3);
       \coordinate (-center) at (.5*#1,.5*#2,-.5*#3);
       \path (0,#2,-.5*#3) coordinate (-left top center) 
                           coordinate (-top left center);
       \path (.5*#1,#2,-#3) coordinate (-top rear center)
                            coordinate (-rear top center);
       \path (#1,#2,-.5*#3) coordinate (-right top center)
                            coordinate (-top right center);
       \path (.5*#1,#2,0) coordinate (-top front center)
                          coordinate (-front top center);
       \path (0,0,-.5*#3) coordinate (-left bottom center) 
                           coordinate (-bottom left center);
       \path (.5*#1,0,-#3) coordinate (-bottom rear center)
                            coordinate (-rear bottom center);
       \path (#1,0,-.5*#3) coordinate (-right bottom center)
                            coordinate (-bottom right center);
       \path (.5*#1,0,0) coordinate (-bottom front center)
                          coordinate (-front bottom center);
       \path (0,.5*#2,0) coordinate (-left front center) 
                           coordinate (-front left center);
       \path (0,.5*#2,-#3) coordinate (-left rear center)
                            coordinate (-rear left center);
       \path (#1,.5*#2,0) coordinate (-right front center)
                            coordinate (-front right center);
       \path (#1,.5*#2,-#3) coordinate (-right rear center)
                          coordinate (-rear right center);
    \end{scope}
}

\tikzset{
  pics/cuboid/.style = {
    setup code = \tikz@lib@cuboid@setup,
    background code = \tikz@lib@cuboid@draw#1\pgf@stop
  },
  pics/cuboid/.default={1--1--1},
  cuboid/.is family,
  cuboid,
  all faces/.style={fill=white},
  all grids/.style={draw=none},
  front face/.style={},
  front grid/.style={},
  right face/.style={},
  right grid/.style={},
  top face/.style={},
  top grid/.style={},
  edges/.style={},
  hidden edges/.style={draw=none},
  xangle/.initial=0,
  yangle/.initial=90,
  zangle/.initial=210,
  xscale/.initial=1,
  yscale/.initial=1,
  zscale/.initial=0.5
}

\newcommand{\tikzcuboidreset}{
\tikzset{cuboid,
  all faces/.style={fill=white},
  all grids/.style={draw=none},
  front face/.style={},
  front grid/.style={},
  right face/.style={},
  right grid/.style={},
  top face/.style={},
  top grid/.style={},
  edges/.style={},
  hidden edges/.style={draw=none},
  xangle=0,
  yangle=90,
  zangle=210,
  xscale=1,
  yscale=1,
  zscale=0.5
}
}
\newcommand{\tikzcuboidset}{\@ifstar\tikzcuboidset@star\tikzcuboidset@nostar} 
\newcommand{\tikzcuboidset@nostar}[1]{\tikzcuboidreset\tikzset{cuboid,#1}}
\newcommand{\tikzcuboidset@star}[1]{\tikzset{cuboid,#1}}

\newcommand\blfootnote[1]{%
  \begingroup
  \renewcommand\thefootnote{}\footnote{#1}%
  \addtocounter{footnote}{-1}%
  \endgroup
}

\makeatother

\title{Estimation of individual causal effects in network setup for multiple treatments}
\author {
    Abhinav Thorat\textsuperscript{\rm 1,\equalcontrib},
    Ravi Kolla\textsuperscript{\rm 1, \equalcontrib},
    Niranjan Pedanekar\textsuperscript{\rm 1},
    Naoyuki Onoe\textsuperscript{\rm 1,\rm 2}
}
\affiliations {
    \textsuperscript{\rm 1}Sony Research India\\
    \textsuperscript{\rm 2}Sony Global Corporation\\
    \texttt{\{abhinav.thorat, ravi.kolla, niranjan.pedanekar, naoyuki.onoe\}@sony.com}
}

\begin{document}
\theoremstyle{definition}
\newtheorem{definition}{Definition}[section]

\maketitle

\begin{abstract}
\blfootnote{This work has been accepted at AAAI-GCLR 2024 workshop.}
We study the problem of estimation of Individual Treatment Effects (ITE) in the context of multiple treatments and networked observational data. Leveraging the network information, we aim to utilize hidden confounders that may not be directly accessible in the observed data, thereby enhancing the practical applicability of the strong ignorability assumption. To achieve this, we first employ Graph Convolutional Networks (GCN) to learn a shared representation of the confounders. Then, our approach utilizes separate neural networks to infer potential outcomes for each treatment. We design a loss function as a weighted combination of two components: representation loss and Mean Squared Error (MSE) loss on the factual outcomes. To measure the representation loss, we extend existing metrics such as Wasserstein and Maximum Mean Discrepancy (MMD) from the binary treatment setting to the multiple treatments scenario. To validate the effectiveness of our proposed methodology, we conduct a series of experiments on the benchmark datasets such as BlogCatalog and Flickr. The experimental results consistently demonstrate the superior performance of our models when compared to baseline methods.
\end{abstract}

\input{introduction}

\input{literature_survey}

\input{problem_formulation}

\input{proposed_architecture}

\input{numerical_results}

\input{conclusion}

\newpage
\bibliography{references}

\end{document}

%% file: introduction.tex
\section{Introduction}
\label{sec:introduction}
In the landscape of causal inference, the estimation of \textit{ITE}s stands as a well-established yet intricate problem. It can be effectively utilized for the dynamic personalization of treatments in various domains, such as economics, education, healthcare, marketing, recommendation systems, and more. For example, it can be used to administer the best drug from the \textit{many available} options \textit{(multiple treatments)} for a patient based on their medical history; implement the best job program from the \textit{many available} choices for a candidate in the education domain; and design the best incentive program from the \textit{many available} options for a customer to prevent churn in any subscription-based industry, such as Over-The-Top (OTT), digital media, telecom, etc. In today's world nearly everyone is interconnected with a subset of others through some form of a \textit{(social) network}. This allows us to harness this connectivity for uncovering hidden information that would have otherwise remained elusive without access to network information.
In this work, we study the problem of the estimation of ITEs in observational studies in the presence of multiple treatments and experimental units are connected through some network. The data we use has been collected or made available in more realistic observational studies, as opposed to the gold standard experiments known as Randomized Control Trials (RCTs), where the data is collected meticulously through a well-designed process. This may not be possible in all use cases due to various reasons such as ethical concerns. To the best of our knowledge, this is the first work on the estimation of ITEs in a networked setup with multiple treatments, which is motivated by the problem setup in~\cite{guo2020learning}.

Strong ignorability is a standard assumption in causal inference literature, essentially stating that all confounders are directly measurable and present in the observed data, with no consideration for hidden confounders. Note that, this assumption is difficult to test in the practice as it requires a deep understanding of the causal relationships among all the variables. For example, in the drug administration use-case from~\cite{guo2020learning}, an individual's socio-economic status serves as a confounding factor that influences both treatment assignment and outcomes. For instance, a lower socio-economic status may hinder one's ability to afford a more expensive drug, potentially leading to a negative impact on their health. However, directly measuring an individual's socio-economic status poses challenges due to the lack of available data. If we do not control for this confounder, it can lead to inaccurate ITE estimates. Consequently, proxies such as age, education, and income are often employed to control for socio-economic status. Nevertheless, in observational studies, the causal relationships between variables remain unknown. Here, we demonstrate that these scenarios can be effectively addressed by leveraging network information to understand the user's community by considering its one-hop neighbours. To be precise, we use the network information to uncover hidden confounders not directly available in the given data and observe superior performance of our models compared to the baselines that do not incorporate network information. Therefore, it is reasonable to argue that the strong ignorability assumption holds better in our case compared to scenarios without network data.     
     
We now briefly highlight the key challenges associated with the problem at hand. It is well known that the observational studies naturally suffer from the treatment selection bias as the assignment of treatments naturally takes place based on the covariates. Additionally, there is a potential issue of observing an imbalance in the number of samples available for each treatment in the given data. We would like to highlight that the above mentioned issues become more dominant especially when dealing with a multiple treatments setup. Furthermore, an individual unit's outcome is causally influenced by the covariates of its one-hop neighbours, which means that the observations are non i.i.d. in nature. We try to counter these challenges by first learning a shared representation of the given units' covariates using GCN~\cite{kipf2016semi}, by leveraging the network information, as the first step of the estimation process. Then, we perform the prediction of counterfactuals through separate neural networks across all treatments by minimizing the weighted sum of loss functions associated with the shared representation and counterfactual prediction. Figure~1 depicts a high-level diagram of our setup and the proposed architecture.
\begin{figure*}
\centering
\resizebox{!}{6cm}{
\begin{tikzpicture}
    \definecolor{lightgray}{HTML}{E2F0D9}
    \definecolor{arrow}{HTML}{000000}
    \definecolor{back}{HTML}{1952A6}
    \definecolor{gcn}{HTML}{BCBEDC}
    \definecolor{phi}{HTML}{DAF467}
    \definecolor{treat}{HTML}{FBA39D}
 	\definecolor{l2}{HTML}{B1D4FC}
	
	\pgfmathsetmacro\squareSize{0.25}
    \pgfmathsetmacro\startX{-2.25}
    \pgfmathsetmacro\startY{-2}
    
    \foreach \x in {1,2,3,4,5,6} {
        \draw[fill=lightgray] (\startX + \x*\squareSize, \startY) rectangle ++(\squareSize, \squareSize);
    }
    
    \pgfmathsetmacro\startX{-2.75}
    \pgfmathsetmacro\startY{-4}
    
    \foreach \x in {1,2,3,4,5,6} {
        \draw[fill=lightgray] (\startX + \x*\squareSize, \startY) rectangle ++(\squareSize, \squareSize);
    }
    
    \pgfmathsetmacro\startX{0.95}
    \pgfmathsetmacro\startY{-2.5}
    
    \foreach \x in {1,2,3,4,5,6} {
        \draw[fill=lightgray] (\startX + \x*\squareSize, \startY) rectangle ++(\squareSize, \squareSize);
    }
    
    \pgfmathsetmacro\startX{2}
    \pgfmathsetmacro\startY{-3.85}
    
    \foreach \x in {1,2,3,4,5,6} {
        \draw[fill=lightgray] (\startX + \x*\squareSize, \startY) rectangle ++(\squareSize, \squareSize);
    }

    \node[businessman, minimum size=0.75cm] (businessman) at (4*\squareSize-1,-2) {};
    \node[charlie, minimum size=0.75cm] (charlie) at (4*\squareSize+2,-3) {};
    \node[alice, minimum size=0.75cm] (alice) at (4*\squareSize-1.5,-4) {};
    \node[dave,female, minimum size=0.75cm] (female) at (4*\squareSize,-3) {};
    
    \draw[darkgray, line width=2pt] (businessman) -- (female);
    \draw[darkgray, line width=2pt] (charlie) -- (female);
    \draw[darkgray, line width=2pt] (alice) -- (female);
    \draw[darkgray, line width=2pt] (alice) -- (charlie);
    
    \node at (1.25, -4.8) [align=center] {Network of users\\ with their covariates};
    
     \draw[-{Latex[length=2.5mm]}, line width=1pt, arrow] (3.75,-3.5) -- (5,-3.5);
     
    \node[draw, rounded corners=2mm, fill=gcn, minimum width=1cm, minimum height=4cm] at (5.5,-3.5) {GCN};

    \node at (5.5, -6.25) [align=center,font=\fontsize{8}{10}] {Graph \\Convolutional\\Layer(s)};

     \draw[-,dashed, line width=1pt, arrow] (6,-3.5) -- (6.75,-3.5);
     \draw[-{Latex[length=2.5mm]}, line width=1pt, arrow] (6.8,-3.5) -- (7.5,-3.5);
     
    \node[draw, rounded corners=2mm, fill=phi, minimum width=1cm, minimum height=4cm,font=\fontsize{14}{18}] at (8,-3.5) {};
	\node [align=center,font=\fontsize{14}{16}]  at (8,-3.5) {$\Phi$};
    \node [align=center,font=\fontsize{8}{10}]  at (8, -6) {Shared \\Representation};

    \node [align=center,font=\fontsize{8}{10}]  at (11.15,-1) {Treatment Head Layer(s)};
	\node[draw, rounded corners=1mm, fill=treat, minimum width=0.75cm, minimum height=1cm] at (9.5,-2) {$f_{0}^0$};
	\node[draw, rounded corners=1mm, fill=treat, minimum width=0.75cm, minimum height=1cm] at (9.5,-3.5) {$f_{0}^K$};

	\node[draw, rounded corners=1mm, fill=treat, minimum width=0.75cm, minimum height=1cm] at (12.75,-2) {$f_{L}^0$};
	\node[draw, rounded corners=1mm, fill=treat, minimum width=0.75cm, minimum height=1cm] at (12.75,-3.5) {$f_{L}^K$};
	
	\node[draw, rounded corners=1mm, minimum width=2.5cm, minimum height=0.70cm] at (10.40,-5) {$\mathcal{L}_2(\Phi(X),T_{obs})$};

	\node [align=center,font=\fontsize{8}{10}]  at (10.40,-5.75) {Representation Loss};
	\node[draw, rounded corners=1mm, fill=lightgray,minimum width=0.75cm, minimum height=2.6cm] at (14.25,-2.75) {};	
	
	\node [align=center,font=\fontsize{12}{14}]  at (14.25,-2) {$\hat{Y}_0$};
	\draw[-,dash dot, line width=1pt, arrow] (14.25,-2.3) -- (14.25,-3.3);
	\node [align=center,font=\fontsize{12}{14}]  at (14.25,-3.5) {$\hat{Y}_K$};
	
	\node[draw, rounded corners=1mm, minimum width=2.5cm, minimum height=0.70cm] at (13.35,-5) {$\mathcal{L}_1(\hat{Y}_{T_{obs}},Y_{obs})$};
	\node [align=center,font=\fontsize{8}{10}]  at (13.35,-5.75) {Regression Loss};
	\draw[-{Latex[length=2.5mm]}, line width=1pt, arrow] (14.25,-4.05)--(14.25,-4.65);
	
	
	\draw[-{Latex[length=2.5mm]}, line width=1pt, arrow] (8.5,-5) -- (9.15,-5);
	
	\draw[-,dashed, line width=1pt, arrow] (9.9,-2) -- (11,-2);
	\draw[-{Latex[length=2.5mm]}, line width=1pt, arrow] (11,-2) -- (12.40,-2);
	
	\draw[-,dashed, line width=1pt, arrow] (9.9,-3.5) -- (11,-3.5);
	\draw[-{Latex[length=2.5mm]},, line width=1pt, arrow] (11,-3.5) -- (12.40,-3.5);
	
    \node (U) at (8.38,-2.75){} ;
	\node (V) at (9.15,-2) {};

	\draw[-{Latex[length=2.5mm]}, line width=1pt,rounded corners=.5pt, arrow]
    (U)
    -- ++(0.35, 0) 
    |- ($(V) + (0, 0)$)
    ;
    
	\draw[-{Latex[length=2.5mm]}, line width=1pt,rounded corners=.5pt, arrow](13.125,-3.5)--(13.9,-3.5);
    
    \node (R) at (8.38,-2.75){} ;
	\node (S) at (9.15,-3.5) {};

	\draw[-{Latex[length=2.5mm]}, line width=1pt,rounded corners=.5pt, arrow]
    (R)
    -- ++(0.35, 0) |- ($(S) + (0, 0.025)$)
    ;   
    
   	\draw[-{Latex[length=2.5mm]}, line width=1pt,rounded corners=.5pt, arrow](13.125,-2)--(13.9,-2);

 \end{tikzpicture}}
\label{fig:workflow-overview}
\caption{A brief overview of the workflow considered in this work.}
\end{figure*}
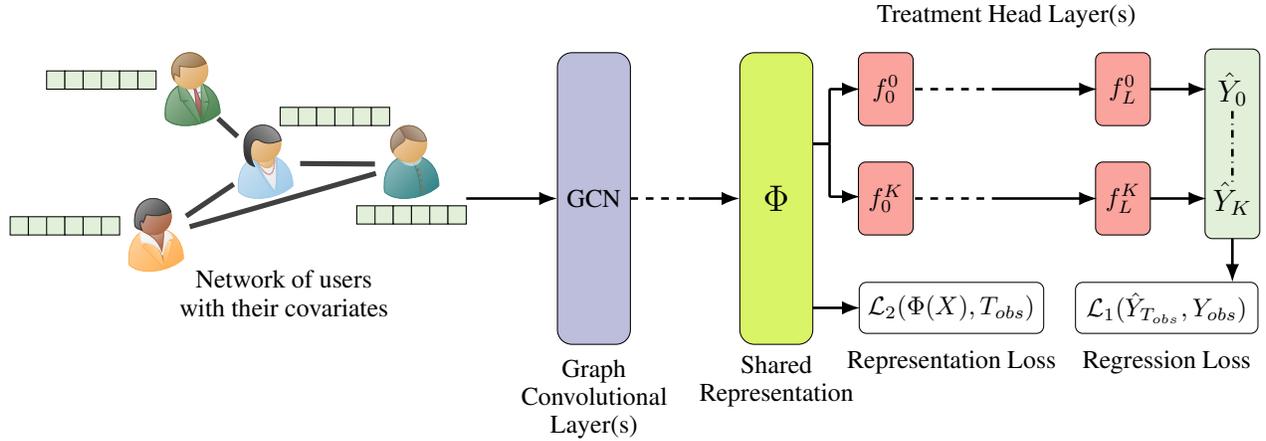

We now briefly present the salient contributions of our work. 
\begin{itemize}
\item We consider a novel problem of the estimation of ITEs in a networked observational studies setting with multiple treatments.
\item  We propose deep learning models that utilize additional network information to mitigate the confounding bias arising from the hidden confounders not directly available in the data.

\item In the context of multiple treatment setting, there is a lack of loss functions to measure the effectiveness of learned representation. We propose an extension of the existing representation loss functions in the binary treatment setting~\cite{shalit2017estimating}, such as Wasserstein~\cite{gretton2012kernel} and MMD~\cite{villani2009optimal, cuturi2014fast}, to suit our framework.

\item We conduct extensive experiments on two standard datasets, BlogCatalog and Flickr~\cite{guo2020learning}, that demonstrate the superior performance of our proposed models compared to the baselines.
\end{itemize} 

We organize the rest of the paper as follows. We provide the details of the related work in the following section. We provide the details of the considered problem statement in the Problem Formulation section. Then, the section Proposed Architecture introduces the architecture of the considered models and their technical details. We provide the details of the considered datasets and comparison of numerical results in the Numerical Results section. We finally conclude the work and provide directions for future work in the Conclusion section. 

%% file: literature_survey.tex
\section{Literature survey}
\label{sec:literature-survey}
Our work deals with the estimation of ITEs, that differs significantly from the estimation of the treatment effects at whole population level such as Average Treatment Effect (ATE)~\cite{shpitser2012identification, pearl2017detecting}. Majority of the ITE estimation works in the literature deals with the case of binary treatments and without the network~\cite{johansson2016learning, wager2018estimation, hill2011bayesian}. However, there have been a limited number of works in the literature that deal with the ITE estimation for multiple treatments without the network setting~\cite{yoon2018ganite, schwab2018perfect, schwab2020learning}. Note that all these works assume that the strong ignorability assumption holds whereas it is difficult to test in the practice and may be violated as well. In contrast, we use the network information to obtain hidden confounders that are not directly available in the observed data that in turn helps the strong ignorability assumption to hold better in practice. We would like to point out that our work has close connections to the work in~\cite{guo2020learning}. Specifically, the authors in~\cite{guo2020learning} consider the ITE estimation task in network setup for binary treatment setting. On the contrary, we extend it to the multiple treatments setting which introduces its own challenges as mentioned above. Furthermore, it is worth mentioning that GCNs have been used in various fields that have network data such as computer vision, natural language processing, traffic, recommender systems, chemistry etc, refer to a survey on GCNs here~\cite{wu2020comprehensive}. Yet, we believe that ours is the first work to consider the application of GCNs in the ITE estimation task under the networked observational studies setting with multiple treatments.     

%% file: problem_formulation.tex
\section{Problem Formulation}
\label{sec:problem-formulation}
In this section, we explain the considered problem statement in detail. We begin with the notation used in this paper. Scalars are denoted by non-bold faced letters, vectors by bold-faced lowercase letters, and matrices by bold-faced uppercase letters. We use subscripts and superscripts to denote instances (units) and treatments, respectively, when both are present in the notation; otherwise, their meaning can be understood from the context. We follow the Rubin-Neyman potential outcomes framework \cite{rubin2005causal} to introduce the problem statement. Let $\mathbf{x} \in \mathbb{R}^p$ and $t \in \mathbb{N}$ denote the covariates and assigned treatment of an instance. We use $K$ to represent the number of available treatments, allowing $K$ to be any finite number greater than or equal to 2. We use $N$ to denote the number of instances in the dataset. Furthermore, let $y_i^t \in \mathbb{R}$ denote the potential outcome of instance $i$ when treatment $t$ is applied. We assume that all instances are connected through a (social) network, with its adjacency matrix denoted by $\mathbf{A}$. It is important to note that in this work, we restrict $\mathbf{A}$ to be undirected and stationary. Additionally, we assume that the outcomes $y_i^t$ of an instance $i$ are causally influenced only by the covariates of its one-hop neighbors.

With this notation, we now define the ITE of treatment-$a$ with respect to treatment $b$ for a given instance-$i$ and the adjacency matrix $\mathbf{A}$ as follows:
\begin{equation}
\label{eq:ite-formula}
\tau^{a, b}(\mathbf{x_i, A}) = \mathbb{E}\left[ y_i^a \mid \mathbf{x_i, A} \right] - \mathbb{E}\left[ y_i^b \mid \mathbf{x_i, A} \right]. 
\end{equation}
Intuitively, for the given instance, ITE measures the uplift obtained by the treatment-$a$ compared to treatment-$b$. Given $N$ instances, their assigned treatments, the corresponding observed (factual) outcomes and the network $\lbrace (\mathbf{x_i}, t_i, y_i^t)_{i=1}^N, \mathbf{A} \rbrace $, our goal is to estimate ITEs of all given instances across all pairs of treatments. Note that the problem is challenging due to the following factors: $(i).$ In observational studies, it is natural to notice the confounding bias resulting from the way treatments are assigned and becomes stronger especially in the multiple treatments setting. $(ii).$ Instances are inherently connected to others through the network, therefore the covariates' correspond to non i.i.d. samples. $(iii).$ The adjacency matrix, $\mathbf{A}$, of the network is potentially high-dimensional and possibly shows substantial sparsity.

For the objective quantification of our proposed models, we extend the standard metrics in the literature such as Rooted \textbf{P}recision in the \textbf{E}stimation of \textbf{H}eterogeneous \textbf{E}ffects, denoted by $\epsilon_{PEHE}$, and Mean Absolute Error on the \textbf{A}verage \textbf{T}reatment \textbf{E}ffect, denoted by $\epsilon_{ATE}$, defined for binary treatment setting~\cite{guo2020learning} to multiple treatment setting by taking the average across all treatment pairs whose formulations are given below. Let $\hat{\tau}^{a, b}(\mathbf{x_i, A})$ denotes the estimate of $\tau^{a, b}(\mathbf{x_i, A})$. Then, 
\begin{equation}
\label{eq:epsilon-pehe-formula} 
\epsilon_{PEHE}  = \frac{1}{{K \choose 2}} \sum\limits_{a=0}^{K-1} \sum\limits_{b=0}^{a-1}\left[ \frac{1}{N} \sum\limits_{i=1}^N (\hat{\tau}^{a, b}(\mathbf{x_i, A}) - \tau^{a, b}(\mathbf{x_i, A}))^2 \right] 
\end{equation}

\begin{multline}
\label{eq:epsilon-ate-formula}
\epsilon_{ATE}  = \frac{1}{{K \choose 2}} \sum\limits_{a=0}^{K-1} \sum\limits_{b=0}^{a-1} \biggl[ \Bigl| \frac{1}{N} \sum\limits_{i=1}^N \hat{\tau}^{a, b}(\mathbf{x_i, A}) \, - \\
\frac{1}{N} \sum\limits_{i=1}^N \tau^{a, b}(\mathbf{x_i, A}) \Bigr| \biggr] 
\end{multline}   

%% file: proposed_architecture.tex
\begin{table*}
\caption{\label{tab:dataset-details}Summary of datasets.}
\begin{tabular} { | c | c | c | c | c | c | c | c |}
\hline
\multirow{2}{*}{Dataset} &  \multirow{2}{*}{\#users(instances)} &  \multirow{2}{*}{\#edges} &  \multirow{2}{*}{\#covariates} & \multirow{2}{*}{$k_2$} &  \multicolumn{3}{|c|}{avg-pairwise-ate} \\
\cline{6-8}
 & & & & & $K=4$ & $K=8$ & $K=16$ \\
\hline
\multirow{3}{*}{BlogCatalog} & \multirow{3}{*}{$5196$} & \multirow{3}{*}{$171743$}  & \multirow{3}{*}{$8189$} & $0.5$ & $ 4.08\pm 0.24$ & $ 2.66 \pm 0.49$ & $2.40 \pm 0.03$ \\
\cline{5-8}
 &  &   &   & 1 & $8.48 \pm 0.38$ & $ 4.89\pm 0.19$ & $ 3.78\pm 0.45$ \\
\cline{5-8}
 &  &   &   & 2 & $ 15.76 \pm 0.85$ & $ 11.73 \pm 1.00$ & $ 6.04 \pm 0.95$ \\
\hline
\multirow{3}{*}{Flickr} & \multirow{3}{*}{$7575$} &  \multirow{3}{*}{$239738$} & \multirow{3}{*}{$12047$} & $0.5$ & $ 2.69 \pm 0.07$ & $1.49 \pm 0.03$ & $1.11 \pm 0.04$ \\
\cline{5-8}
 & & & & $1$ & $4.08 \pm 0.02$ & $2.24 \pm 0.06$ & $1.52 \pm 0.04$ \\
\cline{5-8}
 & & & & $2$ & $ 7.82 \pm 0.08$ & $4.82 \pm 0.04$ & $2.87 \pm 0.02$ \\
\hline
\end{tabular}
\end{table*}

\section{Proposed Architecture}
\label{proposed-architecture}
We begin this section by providing a formal definition to the strong ignorability assumption. 

\textbf{Definition.} Strong ignorability assumption rests on the following two key assumptions. First, it assumes the conditional independence of potential outcomes and treatment assignment given the covariates. In other words it assumes that there are no hidden confounders in the observed data. Second, the probability of receiving any treatment falls \textit{strictly} between $0$ and $1$ for any instance. Mathematically, it is expressed as $\left( y_0, y_1, \ldots y_K \right) \perp t \mid \mathbf{x}$ and $0<P(t=a|\mathbf{x})<1$ $\forall \, \mathbf{x}$ and $\forall \, a \in \{0, 1, \ldots, K \}.$

This assumption leads to the following equality $\mathbb{E}[y_a|\mathbf{x}]=\mathbb{E}[y|\mathbf{x}, t=a]$, where $y_a$ represents the outcome associated with the covariates $\mathbf{x}$ when the treatment-$a$ is applied. Intuitively, it implies that all the confounders are present in the given data. However, the same may not hold in practice as there is a significant chance of existence of hidden confounders that are not directly available in the given data. In that case, the above equality will not hold thus resulting into a biased estimator. Therefore, in our setup, we use the additional network information to obtain any hidden confounders such that the strong ignorability assumption holds in a more realistic manner. 

Our proposed model primarily involves two steps. First step is designed to learn a balanced representation of confounders across all treatments by taking $\mathbf{X}$ and $\mathbf{A}$ as inputs. The second step is focussed on training a separate model for each treatment to predict counterfactuals. A detailed explanation of how our model accomplishes these two steps is given in the following sections.  

\textbf{A. Learning Shared Representation of Confounders with Covariates and Network.}

To mitigate the bias resulted from the assignment of treatments in observational studies we resort to learning a representation of confounders that is balanced across all treatment groups.  
It is to be noted that the representational learning has been utilized for the estimation of causal effects for binary treatments without the network in the literature~\cite{johansson2016learning, shalit2017estimating}. To learn representations from auxiliary network information and observational data, we refer to \cite{guo2020learning}, where the authors utilized the GCN for learning representations from the network and observational data for a binary treatment setup. Therefore, in the first part of our model, the representation learning function, $\Phi: \mathbf{X} \times \mathbf{A} \rightarrow \mathbb{R}^d,$ is parameterized using a GCN with multiple layers, whose effectiveness has been verified in various machine learning tasks across different types of networked data~\cite{wu2020comprehensive}. Mathematical formulation of $\Phi$ for a single GCN layer~\footnote{To keep the notational complexity simple we provided the formulation for single GCN layer here. However, the same can be extended to any number of GCN layers as well.} is as follows: $\Phi(\mathbf{X, A}) = \sigma(\mathbf{(A+ I_n)XU})$, where $\mathbf{U} \in \mathbb{R}^{n \times d}$ represents the weight matrix to be learned, and $\sigma$ stands for the ELU activation function. The challenge here lies in using an appropriate loss function for measuring the effectiveness of a representation in multiple treatments setting which will be discussed in the further sections.    

\textbf{B. Potential Outcomes Prediction.}
The second part of the model is dedicated for predicting potential outcomes of all treatments using the obtained shared representation detailed in the above section, and observed treatment. As there are $K$ available treatments, we use $f^t: \mathbb{R}^d \times \{t\} \rightarrow \mathbb{R}$ to denote the potential outcome prediction function of the treatment-$t.$ We use $L$ number of Fully Connected (FC) layers followed by a regression layer for learning each function $f^t$ that we refer as the network-head of treatment-$t.$ Hence, in this part, we have an overall of $K$ number of network heads corresponding to $K$ distinct treatments. Let $\Phi_i \in \mathbb{R}^d$ be the learned shared representation of confounders of instance-$i$. Then, mathematically, $f^t$ is defined as $f^t = w^t \sigma(W_L^{t} \ldots \sigma(W_1^{t},\Phi_i))$ for an instance-$i$ where $W_l^{t}$ and $w^{t}$ denote the respective weights of the $l^{th}$ FC layer and the regression layer in the network-head-$t$. The equations of neural network bias terms follow the same rule and are omitted here due to notational complexity. With both parts of the model described, our model's prediction of potential outcome of a treatment-$t$ for the given instance-$i$ is defined as $\hat{y}_i^t = f^t(\Phi_i = \Phi(\mathbf{X, A})_i).$

\textbf{C. Loss Function.}
Our proposed model optimizes on a combination of two loss functions namely Regression Loss and Representation Loss across treatments. \\
\textit{(i). Regression Loss.} This is a traditional loss function which is responsible for obtaining a good predictive accuracy of a model. It is the mean squared error between the ground truth, $y_i$, and the predicted potential outcomes, $\hat{y}_i^t \, \forall i, t$ which is given as: 
\begin{equation}
\mathcal{L}_1 = \frac{1}{N} \sum_{i=0}^N (\hat{y}_i^t- y_i)^{2}. 
\end{equation}
In the case of observational studies where each treatment group potentially exhibits a distinct distribution, minimizing the regression loss alone is insufficient for obtaining better models. Therefore, we consider another loss function that aims to measure the balance of covariates' representation distributions across all treatments. \\
\textit{(ii). Representation Loss.} 
Recall that the first part of our model's goal is to obtain a balanced representation of covariates across all treatments to mitigate the treatment assignment bias. Hence, we look for a metric that measures the effectiveness of a balanced representation. The authors in~\cite{shalit2017estimating} use Integral Probability Metrics (IPM) for this task in the context of binary treatments. In particular, they use special cases of IPM such as Wasserstein and Maximum Mean Discrepancy (MMD) metrics for measuring the distance between the treated and control covariates' distributions. We use $\Pi$ to denote these metrics that take two distributions as inputs. In this work we propose extensions of these metrics to our multiple treatments setting by taking the average of the Wasserstein/MMD metrics computed between distributions corresponding to all pairwise combinations of treatments. Mathematically it is given as below:
\begin{equation}
\mathcal{L}_2 = \frac{1}{\binom{K}{2}} \sum_{a=0}^{K-1} \sum_{b=0}^{a-1} \Pi \left( \lbrace \Phi(\mathbf{X, A}) \rbrace_{t=a}, \lbrace \Phi(\mathbf{X, A}) \rbrace_{t=b} \right),
\end{equation}
where $\lbrace \Phi(\mathbf{X, A}) \rbrace_{t=a}$ and $\lbrace \Phi(\mathbf{X, A}) \rbrace_{t=b}$ denote the representation $\Phi(\mathbf{X, A})$ restricted to samples that received treatment-$a$ and $b$ respectively. Intuitively, the minimization of the above metric results into a representation that is balanced across all pairwise treatments. We try to balance both the above loss functions, $\mathcal{L}_1$ and $\mathcal{L}_2$ for improved models by taking a weighted sum of them which is given below:
\begin{multline}
\label{eq:total-loss}
\mathcal{L} = \alpha \mathcal{L}_1 + \beta \mathcal{L}_2  = \alpha \cdot \frac{1}{N} \sum_{i=0}^{N} (\hat{y}_i^{t} - y_i)^{2} + \\ \beta \cdot \frac{1}{\binom{K}{2}} \sum_{a=0}^{K-1} \sum_{b=0}^{a-1} \Pi \left( \lbrace \Phi(\mathbf{X, A}) \rbrace_{t=a}, \lbrace \Phi(\mathbf{X, A}) \rbrace_{t=b} \right),
\end{multline}
where $\alpha>0$ and $\beta > 0$ are the chosen weights. We propose two  model variants denoted as GCN-Wass and GCN-MMD obtained by choosing $\Pi$ in the above equation to Wasserstein and MMD metrics, respectively.  

%% file: numerical_results.tex
\begin{table*}
\caption{\label{tab:blogcatalog-k-4}Results comparison on BlogCatalog dataset and the number of treatments, $K$, equals to 4.}
\begin{tabular} { | c | c | c | c | c | c | c |}
\hline
\multicolumn{7}{| c |}{BlogCatalog for $K=4$} \\
\hline
\multicolumn{1} { | c  }{$k_2$} & \multicolumn{2}{| c }{0.5} & \multicolumn{2}{| c }{1} & \multicolumn{2}{| c |}{2}\\
\hline
& $\sqrt{\epsilon_{PEHE}}$ & $\epsilon_{ATE}$ & $\sqrt{\epsilon_{PEHE}}$ & $\epsilon_{ATE}$ & $\sqrt{\epsilon_{PEHE}}$ & $\epsilon_{ATE}$\\
\hline
GCN-Wass & $\mathbf{4.95 \pm 1.00}$ & $\mathbf{1.11 \pm 0.62}$ & $\mathbf{7.16 \pm 1.51}$ & $\mathbf{1.77 \pm 0.82}$ & $\mathbf{12.98 \pm 2.72}$ & $\mathbf{3.79 \pm 1.60}$ \\
\hline
GCN-MMD & $5.01 \pm 0.93$  & $1.24 \pm 0.63$  & $7.47 \pm 2.21$  & $2.16 \pm 1.65$  & $13.63 \pm 3.11$  & $4.48 \pm 1.92$\\
\hline
TARNet & $5.52 \pm 1.62$ & $2.03 \pm 1.22$ & $9.96 \pm 2.77$  & $4.28 \pm 1.94$  & $20.43 \pm 5.55$ & $10.38 \pm 3.73$ \\
\hline
CFRNet-Wass & $5.48 \pm 1.62$  & $2.11 \pm 1.15$ & $10.36 \pm 2.99$  & $4.74 \pm 2.12$  & $20.12 \pm 5.79$  & $10.1 \pm 4.26$\\
\hline
CFRNet-MMD & $5.51 \pm 1.59$  & $2.02 \pm 1.23$ & $10.38 \pm 2.95$  & $4.72 \pm 2.08$ & $20.49 \pm 5.56$ & $10.30 \pm 3.52$ \\
\hline
\end{tabular}
\end{table*}

\begin{table*}
\caption{\label{tab:flickr-k-4}Results comparison on Flickr dataset and the number of treatments, $K$, equals to 4.}
\begin{tabular} { | c | c | c | c | c | c | c |}
\hline
\multicolumn{7}{| c |}{Flickr for $K=4$} \\
\hline
\multicolumn{1} { | c  }{$k_2$} & \multicolumn{2}{| c }{0.5} & \multicolumn{2}{| c }{1} & \multicolumn{2}{| c |}{2}\\
\hline
 & $\sqrt{\epsilon_{PEHE}}$ & $\epsilon_{ATE}$ & $\sqrt{\epsilon_{PEHE}}$ & $\epsilon_{ATE}$ & $\sqrt{\epsilon_{PEHE}}$ & $\epsilon_{ATE}$\\
\hline
GCN-Wass & $4.25 \pm 0.54$ & $0.88 \pm 0.25$ & $\mathbf{5.84 \pm 1.02}$ & $\mathbf{1.26 \pm 0.45}$ & $11.67 \pm 3.37$ & $2.64 \pm 1.24$ \\
\hline
GCN-MMD & $\mathbf{4.25 \pm 0.50}$  & $\mathbf{0.89 \pm 0.19}$  & $5.95 \pm 0.99$  & $1.38 \pm 0.39$  & $\mathbf{11.48 \pm 2.81}$  & $\mathbf{2.55 \pm 0.54}$\\
\hline
TARNet & $6.47 \pm 0.68$ & $1.97 \pm 0.81$ & $13.09 \pm 1.78$  & $5.19 \pm 1.96$  & $27.79 \pm 3.18$ & $12.50 \pm 3.29$ \\
\hline
CFRNet-Wass & $6.51 \pm 0.72$  & $2.06 \pm 0.75$ & $12.82 \pm 1.76$  & $5.09 \pm 1.81$  & $26.90 \pm 3.20$  & $12.61 \pm 3.30$\\
\hline
CFRNet-MMD & $6.51 \pm 0.67$  & $1.98 \pm 0.85$ & $13.20 \pm 1.77$  & $5.20 \pm 1.95$ & $27.72 \pm 3.18$ & $12.52 \pm 3.20$ \\
\hline
\end{tabular}
\end{table*}

\begin{table*}
\caption{\label{tab:blogcatalog-k-8}Results comparison on BlogCatalog dataset and the number of treatments, $K$, equals to 8.}
\begin{tabular} { | c | c | c | c | c | c | c |}
\hline
\multicolumn{7}{| c |}{BlogCatlog for $K=8$} \\
\hline 
\multicolumn{1} { | c  }{$k_2$} & \multicolumn{2}{| c }{0.5} & \multicolumn{2}{| c }{1} & \multicolumn{2}{| c |}{2}\\
\hline
 & $\sqrt{\epsilon_{PEHE}}$ & $\epsilon_{ATE}$ & $\sqrt{\epsilon_{PEHE}}$ & $\epsilon_{ATE}$ & $\sqrt{\epsilon_{PEHE}}$ & $\epsilon_{ATE}$\\
\hline
GCN-Wass & $6.86 \pm 1.50$ & $\mathbf{2.34 \pm 0.65}$ & $\mathbf{11.29 \pm 3.42}$ & $\mathbf{4.14 \pm 1.40}$ & $\mathbf{23.49 \pm 5.17}$ & $\mathbf{8.91 \pm 2.20}$ \\
\hline
GCN-MMD & $6.97 \pm 1.58$  & $2.48 \pm 0.71$  & $11.51 \pm 3.38$  & $4.38 \pm 1.46$  & $24.67 \pm 5.52$  & $10.02 \pm 2.86$\\
\hline
TARNet & $6.66 \pm 1.53$ & $2.49 \pm 0.70$ & $12.70 \pm 2.51$  & $6.00 \pm 1.21$  & $24.22 \pm 4.45$ & $12.52 \pm 2.17$ \\
\hline
CFRNet-Wass & $\mathbf{6.53 \pm 1.45}$  & $2.46 \pm 0.64$ & $12.55 \pm 2.64$  & $5.99 \pm 1.29$  & $24.69 \pm 4.39$  & $13.33 \pm 2.37$\\
\hline
CFRNet-MMD & $7.80 \pm 1.91$  & $2.46 \pm 0.68$ & $12.64 \pm 2.74$  & $6.03 \pm 1.29$ & $24.58 \pm 4.78$ & $12.88 \pm 2.45$ \\
\hline
\end{tabular}
\end{table*}

\begin{table*}
\caption{\label{tab:flickr-k-8}Results comparison on Flickr dataset and the number of treatments, $K$, equals to 8.}
\begin{tabular} { | c | c | c | c | c | c | c |}
\hline
\multicolumn{7}{| c |}{Flickr for $K=8$} \\
\hline 
\multicolumn{1} { | c  }{$k_2$} & \multicolumn{2}{| c }{0.5} & \multicolumn{2}{| c }{1} & \multicolumn{2}{| c |}{2}\\
\hline
 & $\sqrt{\epsilon_{PEHE}}$ & $\epsilon_{ATE}$ & $\sqrt{\epsilon_{PEHE}}$ & $\epsilon_{ATE}$ & $\sqrt{\epsilon_{PEHE}}$ & $\epsilon_{ATE}$\\
\hline
GCN-Wass & $4.86 \pm 0.70$ & $1.51 \pm 0.33$ & $\mathbf{7.63 \pm 0.97}$ & $\mathbf{1.84 \pm 0.38}$ & $\mathbf{16.69 \pm 2.99}$ & $\mathbf{3.53 \pm 1.00}$ \\
\hline
GCN-MMD & $\mathbf{4.86 \pm 0.69}$  & $\mathbf{1.50 \pm 0.31}$  & $7.70 \pm 1.04$  & $1.86 \pm 0.43$  & $16.87 \pm 3.19$  & $3.61 \pm 1.08$\\
\hline
TARNet & $6.44 \pm 0.56$ & $2.49 \pm 0.70$ & $13.94 \pm 1.83$  & $5.84 \pm 1.22$  & $29.18 \pm 3.82$ & $14.58 \pm 2.50$ \\
\hline
CFRNet-Wass & $6.22 \pm 0.60$  & $2.06 \pm 0.45$ & $13.20 \pm 1.52$  & $5.69 \pm 1.29$  & $28.20 \pm 3.15$  & $14.53 \pm 2.42$\\
\hline
CFRNet-MMD & $6.45 \pm 0.63$  & $2.05 \pm 0.44$ & $13.96 \pm 1.83$  & $5.80 \pm 1.38$ & $29.26 \pm 3.82$ & $14.65 \pm 2.44$ \\
\hline
\end{tabular}
\end{table*}

\begin{table*}
\caption{\label{tab:blogcatalog-k-16}Results comparison on BlogCatalog dataset and the number of treatments, $K$, equals to 16.}
\begin{tabular} { | c | c | c | c | c | c | c |}
\hline
\multicolumn{7}{| c |}{BlogCatlog for $K=16$} \\
\hline 
\multicolumn{1} { | c  }{$k_2$} & \multicolumn{2}{| c }{0.5} & \multicolumn{2}{| c }{1} & \multicolumn{2}{| c |}{2}\\
\hline
 & $\sqrt{\epsilon_{PEHE}}$ & $\epsilon_{ATE}$ & $\sqrt{\epsilon_{PEHE}}$ & $\epsilon_{ATE}$ & $\sqrt{\epsilon_{PEHE}}$ & $\epsilon_{ATE}$\\
\hline
GCN-Wass & $7.53 \pm 1.12$ & $\mathbf{2.46 \pm 0.62}$ & $\mathbf{12.00 \pm 2.00}$ & $\mathbf{4.39 \pm 1.20}$ & $\mathbf{21.90 \pm 4.00}$ & $\mathbf{8.42 \pm 2.33}$ \\
\hline
GCN-MMD & $9.09 \pm 1.73$  & $3.68 \pm 1.10$  & $15.68 \pm 3.00$  & $5.89 \pm 1.67$  & $30.34 \pm 4.50$  & $11.40 \pm 1.95$\\
\hline
TARNet & $7.07 \pm 1.46$ & $2.72 \pm 1.00$ & $13.45 \pm 2.95$  & $6.84 \pm 1.90$  & $26.25 \pm 5.04$ & $14.72 \pm 2.90$ \\
\hline
CFRNet-Wass & $\mathbf{7.02 \pm 1.48}$  & $2.78 \pm 0.96$ & $13.25 \pm 2.95$  & $6.80 \pm 1.95$  & $26.42 \pm 5.00$  & $14.91 \pm 2.88$\\
\hline
CFRNet-MMD & $7.08 \pm 1.51$  & $2.74 \pm 1.02$ & $13.49 \pm 2.73$  & $7.03 \pm 1.69$ & $26.36 \pm 4.66$ & $14.94 \pm 2.49$ \\
\hline
\end{tabular}
\end{table*}

\begin{table*}
\caption{\label{tab:flickr-k-16}Results comparison on Flickr dataset and the number of treatments, $K$, equals to 16.}
\begin{tabular} { | c | c | c | c | c | c | c |}
\hline
\multicolumn{7}{| c |}{Flickr for $K=16$} \\
\hline 
\multicolumn{1} { | c  }{$k_2$} & \multicolumn{2}{| c }{0.5} & \multicolumn{2}{| c }{1} & \multicolumn{2}{| c |}{2}\\
\hline
 & $\sqrt{\epsilon_{PEHE}}$ & $\epsilon_{ATE}$ & $\sqrt{\epsilon_{PEHE}}$ & $\epsilon_{ATE}$ & $\sqrt{\epsilon_{PEHE}}$ & $\epsilon_{ATE}$\\
\hline
GCN-Wass & $\mathbf{6.47 \pm 0.32}$ & $\mathbf{0.82 \pm 0.11}$ & $\mathbf{11.06 \pm 0.64}$ & $\mathbf{1.43 \pm 0.18}$ & $\mathbf{20.65 \pm 1.12}$ & $\mathbf{2.76 \pm 0.34}$ \\
\hline
GCN-MMD & $6.59 \pm 0.53$  & $2.32 \pm 0.33$  & $13.50 \pm 2.37$  & $3.48 \pm 0.43$  & $33.24 \pm 6.11$  & $7.93 \pm 1.32$\\
\hline
TARNet & $8.28 \pm 0.79$ & $2.66 \pm 0.49$ & $16.19 \pm 1.36$  & $7.08 \pm 0.72$  & $31.13 \pm 2.62$ & $16.06 \pm 1.61$ \\
\hline
CFRNet-Wass & $8.08 \pm 0.89$  & $2.65 \pm 0.50$ & $15.89 \pm 1.46$  & $7.03 \pm 8.81$  & $30.29 \pm 2.49$  & $16.06 \pm 1.54$\\
\hline
CFRNet-MMD & $8.40 \pm 0.84$  & $2.66 \pm 0.50$ & $16.10 \pm 1.48$  & $6.99 \pm 0.72$ & $31.10 \pm 2.88$ & $16.08 \pm 1.59$ \\
\hline
\end{tabular}
\end{table*}
\section{Numerical Results}
\label{sec:numerical-results}
Evaluation of causal effect models is challenging in practice as there is no available ground truth data. Due to that, it has become a convention in the literature to use either synthetic or semi-synthetic datasets for the evaluation. In this work, we use two standard datasets, BlogCatalog and Flickr used in~\cite{guo2020learning}, to evaluate our proposed models. These datasets are semi-synthetic implying that the covariates and the adjacency matrix correspond to the real world data, while others are synthetically generated. It is important to note that these datasets are used in the literature under binary treatments setting. Here, we extend the same to multiple treatments scenario as detailed in the following subsection. 
\subsection{Datasets generation}
\label{sec:datasets-generation}
\textbf{BlogCatalog:} It is an online platform that hosts blogs posted by users. An instance (a row) in the dataset corresponds to a blogger, and the covariates are a bag-of-words representation of the blogger's keywords in the description. The graph represents the social network among the bloggers. Here, the treatments correspond to devices such as type of mobile (Android or iOS), tablet, computer, etc. used for consuming the blogs, and the outcomes correspond to the readers' opinion of the blogs. \\
\textbf{Flickr:} It is an online platform for sharing images and videos among users connected through a network. Each instance in the dataset corresponds to a user, and the covariates consist of a list of tags of interest. Similar to the BlogCatalog dataset, treatments represent devices, and outcomes represent users' opinion on the shared content.

To generate the potential outcomes of all available treatments, we first generate $K$ centroids as follows. We train an LDA topic model on a large set of documents with a fixed number of topics. Let $\mathbf{z(x_i)}$ denote the topic distribution of unit-$i$. Then, we assign $(K-1)$ centroids as the topic distributions of randomly selected $(K-1)$ units and the $K^{th}$ centroid as the mean of the topic distributions of all the units. We use $\mathbf{z_i}$ to denote the centroid-$i$.  With this notation, we define unscaled potential outcomes $p^i_a$ for unit-$i$ and treatment-$a$ as follows:      
\begin{equation}
\label{eq:unscaled-potential-outcomes}
p^i_a = k_1 \mathbf{z(x_i)^T z_a} + k_2 \sum_{j \in \mathcal{N}(i)} \mathbf{z(x_j)^T z_a},
\end{equation}
where $k_1, k_2 \geq 0$ capture the effect of confounding bias on outcomes of unit-$i$ through itself and its immediate neighbours, respectively. The treatment assignment probabilities of a unit-$i$, $\mathbb{P}(t=a | \mathbf{x_i}, \mathbf{A}),$ are given below: 
\begin{equation}
\label{eq:treatment-assignment-probability}
\mathbb{P}(t=a | \mathbf{x_i}, \mathbf{A}) = \frac{\exp \left( p^i_a \right)}{\sum\limits_{a'=0}^K \exp \left( p^i_{a'} \right) }. 
\end{equation}
Then, the final potential outcomes are given as follows:
\begin{equation}
\label{eq:scaled-potential-outcomes}
y^i_t = 
\begin{cases}
Cp^i_0 + \epsilon, & \text{if } t=0 \\
C \left[ p^i_0 + \sum\limits_{a=1}^K \mathbb{I}(t=a) p^i_a \right] + \epsilon, & \text{if t $\in \lbrace 1, 2, \dots, K \rbrace$},
\end{cases}
\end{equation}  
where $\mathbb{I}(\cdot)$, $C$ and $\epsilon$ are Indicator function, a constant and a random variable drawn from $\mathcal{N}(0, 1)$, respectively. We provide a summary of the above generated datasets in Table~\ref{tab:dataset-details} where avg-pairwise-ate contains the mean and standard deviation of the average of ATEs computed using all pair combinations of treatments across 10 random runs.  
\subsection{Baselines and Results Comparison}
It is important to note that, to the best of our knowledge, there are no prior works in the literature addressing the estimation of ITEs in the context of network data with multiple treatments. Therefore, we naturally chose to extend models from the binary treatment setting, such as TARNET, CFRNET-Wass, and CFRNET-MMD proposed in~\cite{shalit2017estimating}, to multiple treatments. These models initially learn a shared representation of the given covariates and subsequently predict outcomes for all treatments using separate head networks. It is worth mentioning that these models are developed under the assumption of strong ignorability. 

Now, we briefly present the settings considered in the experiments and the obtained results. We conduct experiments for various of number of treatments, $K,$ such as $4, 8$ and $16.$ As the hidden confounding bias increases with the increase of the parameter $k_2$ in equation~\eqref{eq:unscaled-potential-outcomes}, we study its behaviour by running experiments for various values of $k_2$ from $\lbrace 0.5, 1, 2 \rbrace.$ We use $50$ number of topics to train an LDA topic model on both the datasets separately for the calculation of centroids.  
We fine-tune the parameters of all our models, GCN-Wass, GCN-MMD and the baselines to achieve their best performances. For our proposed models, we use 3 GCN layers with 25 nodes each for the representation learning part and 2 Fully Connected (FC) layers with 10 nodes each for the counterfactual prediction of each treatment. For the baselines, we use 2 FC layers with 25 nodes each in the representation stage and 2 FC layers with 25 nodes each for the counterfactual prediction. We employ a batch size of 512 and apply an $L2-$ regularizer with a weight of $0.01.$ We use a learning rate of $0.01$ for all the models. We use the weights of loss functions, $\alpha$ and $\beta$, given in the equation~\eqref{eq:total-loss} as $1$ and $0.5$ respectively. We choose the value of $C$ as $5$ in the equation~\eqref{eq:scaled-potential-outcomes}. We repeat experiments for 10 iterations and report the mean and standard deviation values in the results given in Table~\ref{tab:blogcatalog-k-4}-\ref{tab:flickr-k-16}. 
Note that we denote the best results using bold-face in the tables. From the results, we observe that our models outperform the baselines by a significant margins on both the datasets and on both the metrics $\sqrt{\epsilon_{PEHE}}$ and $\epsilon_{ATE}$ for all values of $k_2.$ Further, note that the gap in the performance of our models to the baselines increases as $k_2$ increases. This confirms that our models significantly improve in performance with the additional component of learning hidden confounders that is in tune with the findings of~\citep{guo2020learning} for the binary treatment setting. 

%% file: conclusion.tex
\section{Conclusion}
\label{sec:conclusion}  
We proposed and studied a challenging yet crucial problem: the estimation of ITEs under networked observational studies with multiple treatments. We utilized network information to mitigate hidden confounding bias, thereby enhancing the practical adherence to the strong ignorability assumption. Our approach used GCNs to construct more sophisticated neural network architectures suitable for ITE estimation. Our numerical results establish the superior performance of our models compared to the baselines. We conclude the work by highlighting a few promising directions for future research. 
While our work exclusively addresses stationary networks, real-world networks are often dynamic in nature. An intriguing avenue for future exploration involves extending ITE estimation techniques to evolving networks, including weighted and directional networks. Additionally, a particularly captivating yet formidable challenge for future research is advancing ITE estimations within this framework to accommodate continuous treatment dosage values.

%% file: gclr-aaai-2024.bbl
\begin{thebibliography}{16}
\providecommand{\natexlab}[1]{#1}

\bibitem[{Cuturi and Doucet(2014)}]{cuturi2014fast}
Cuturi, M.; and Doucet, A. 2014.
\newblock Fast computation of Wasserstein barycenters.
\newblock In \emph{International conference on machine learning}, 685--693.
  PMLR.

\bibitem[{Gretton et~al.(2012)Gretton, Borgwardt, Rasch, Sch{\"o}lkopf, and
  Smola}]{gretton2012kernel}
Gretton, A.; Borgwardt, K.~M.; Rasch, M.~J.; Sch{\"o}lkopf, B.; and Smola, A.
  2012.
\newblock A kernel two-sample test.
\newblock \emph{The Journal of Machine Learning Research}, 13(1): 723--773.

\bibitem[{Guo, Li, and Liu(2020)}]{guo2020learning}
Guo, R.; Li, J.; and Liu, H. 2020.
\newblock Learning individual causal effects from networked observational data.
\newblock In \emph{Proceedings of the 13th international conference on web
  search and data mining}, 232--240.

\bibitem[{Hill(2011)}]{hill2011bayesian}
Hill, J.~L. 2011.
\newblock Bayesian nonparametric modeling for causal inference.
\newblock \emph{Journal of Computational and Graphical Statistics}, 20(1):
  217--240.

\bibitem[{Johansson, Shalit, and Sontag(2016)}]{johansson2016learning}
Johansson, F.; Shalit, U.; and Sontag, D. 2016.
\newblock Learning representations for counterfactual inference.
\newblock In \emph{International conference on machine learning}, 3020--3029.
  PMLR.

\bibitem[{Kipf and Welling(2016)}]{kipf2016semi}
Kipf, T.~N.; and Welling, M. 2016.
\newblock Semi-supervised classification with graph convolutional networks.
\newblock \emph{arXiv preprint arXiv:1609.02907}.

\bibitem[{Pearl(2017)}]{pearl2017detecting}
Pearl, J. 2017.
\newblock Detecting Latent Heterogeneity.
\newblock \emph{Sociological Methods \& Research}, 46(3): 370--389.

\bibitem[{Rubin(2005)}]{rubin2005causal}
Rubin, D.~B. 2005.
\newblock Causal inference using potential outcomes: Design, modeling,
  decisions.
\newblock \emph{Journal of the American Statistical Association}, 100(469):
  322--331.

\bibitem[{Schwab et~al.(2020)Schwab, Linhardt, Bauer, Buhmann, and
  Karlen}]{schwab2020learning}
Schwab, P.; Linhardt, L.; Bauer, S.; Buhmann, J.~M.; and Karlen, W. 2020.
\newblock Learning counterfactual representations for estimating individual
  dose-response curves.
\newblock In \emph{Proceedings of the AAAI Conference on Artificial
  Intelligence}, volume~34, 5612--5619.

\bibitem[{Schwab, Linhardt, and Karlen(2018)}]{schwab2018perfect}
Schwab, P.; Linhardt, L.; and Karlen, W. 2018.
\newblock Perfect match: A simple method for learning representations for
  counterfactual inference with neural networks.
\newblock \emph{arXiv preprint arXiv:1810.00656}.

\bibitem[{Shalit, Johansson, and Sontag(2017)}]{shalit2017estimating}
Shalit, U.; Johansson, F.~D.; and Sontag, D. 2017.
\newblock Estimating individual treatment effect: generalization bounds and
  algorithms.
\newblock In \emph{International conference on machine learning}, 3076--3085.
  PMLR.

\bibitem[{Shpitser and Pearl(2012)}]{shpitser2012identification}
Shpitser, I.; and Pearl, J. 2012.
\newblock Identification of conditional interventional distributions.
\newblock \emph{arXiv preprint arXiv:1206.6876}.

\bibitem[{Villani et~al.(2009)}]{villani2009optimal}
Villani, C.; et~al. 2009.
\newblock \emph{Optimal transport: old and new}, volume 338.
\newblock Springer.

\bibitem[{Wager and Athey(2018)}]{wager2018estimation}
Wager, S.; and Athey, S. 2018.
\newblock Estimation and inference of heterogeneous treatment effects using
  random forests.
\newblock \emph{Journal of the American Statistical Association}, 113(523):
  1228--1242.

\bibitem[{Wu et~al.(2020)Wu, Pan, Chen, Long, Zhang, and
  Philip}]{wu2020comprehensive}
Wu, Z.; Pan, S.; Chen, F.; Long, G.; Zhang, C.; and Philip, S.~Y. 2020.
\newblock A comprehensive survey on graph neural networks.
\newblock \emph{IEEE transactions on neural networks and learning systems},
  32(1): 4--24.

\bibitem[{Yoon, Jordon, and Van Der~Schaar(2018)}]{yoon2018ganite}
Yoon, J.; Jordon, J.; and Van Der~Schaar, M. 2018.
\newblock GANITE: Estimation of individualized treatment effects using
  generative adversarial nets.
\newblock In \emph{International conference on learning representations}.

\end{thebibliography}
